\documentclass[preprint]{elsarticle}
\usepackage[utf8]{inputenc}
\usepackage[margin=4cm]{geometry}
\usepackage{adjustbox}
\usepackage{amssymb}
\usepackage{amsmath,amsthm}
\usepackage{amsfonts}
\usepackage{enumitem}
\usepackage[ruled,vlined]{algorithm2e}
\usepackage{times}
\usepackage{array}
\usepackage{comment}
\usepackage{hyperref}
\usepackage{url}
\usepackage[usenames,dvipsnames]{xcolor}
\usepackage{pdfpages}
\usepackage{footnote}
\usepackage{array,multirow,varwidth}
\usepackage{booktabs}
\usepackage{textcomp}
\usepackage{makecell}
\usepackage{moresize}
\usepackage{fontawesome5}
\usepackage[flushleft]{threeparttable}

\def\arraystretch{1.3}
\journal{Information Processing and Management}
\begin{document}
\begin{frontmatter}
\title{Improving Transfer Learning for Movie Trailer Genre Classification using a Dual Image and Video Transformer}

  \author[mymainaddress]{Ricardo Montalvo-Lezama\corref{mycorrespondingauthor}}
  \cortext[mycorrespondingauthor]{Corresponding author}
  \ead{ricardoml@turing.iimas.unam.mx}

  \author[mymainaddress]{Berenice Montalvo-Lezama}
  \ead{bereml@turing.iimas.unam.mx}

  \author[mymainaddress]{Gibran Fuentes-Pineda}
  \ead{gibranfp@unam.mx}

  \address[mymainaddress]{Instituto de Investigaciones en Matemáticas Aplicadas y en Sistemas, Universidad Nacional Autónoma de México, Circuito Escolar s/n 4to piso, Ciudad Universitaria, Coyoacán, 04510, CDMX, México}

\begin{abstract}
In this paper, we study the transferability of ImageNet spatial and Kinetics spatio-temporal representations to multi-label Movie Trailer Genre Classification (MTGC).
In particular, we present an extensive evaluation of the transferability of ConvNet and Transformer models pretrained on ImageNet and Kinetics to Trailers12k, a new manually-curated movie trailer dataset composed of 12,000 videos labeled with 10 different genres and associated metadata.
We analyze different aspects that can influence transferability, such as frame rate, input video extension, and spatio-temporal modeling.
In order to reduce the spatio-temporal structure gap between ImageNet/Kinetics and Trailers12k, we propose Dual Image and Video Transformer Architecture (DIViTA), which performs shot detection so as to segment the trailer into highly correlated clips, providing a more cohesive input for pretrained backbones and improving transferability (a 1.83\% increase for ImageNet and 3.75\% for Kinetics).
Our results demonstrate that representations learned on either ImageNet or Kinetics are comparatively transferable to Trailers12k. Moreover, both datasets provide complementary information that can be combined to improve classification performance (a 2.91\% gain compared to the top single pretraining).
Interestingly, using lightweight ConvNets as pretrained backbones resulted in only a 3.46\% drop in classification performance compared with the top Transformer while requiring only 11.82\% of its parameters and 0.81\% of its FLOPS.
\end{abstract}

\end{frontmatter}

\section{Introduction}
\label{sec:intro}

The success of Transfer Learning (TL) has been mainly driven by the availability of large and diverse source datasets.
For image analysis tasks, image classification (IC) on ImageNet~\citep{imagenet} is a standard pretraining practice.
Similarly, ImageNet pretraining has been commonly leveraged to initialize video analysis models, especially for human action recognition (HAR).
On the other hand, human action recognition has been the main video pretraining task for TL on video analysis tasks.
As opposed to the image domain, in the video domain different datasets (e.g., UCF~\citep{ucf101}, Kinetics~\citep{kinetics400}, or AVA~\citep{ava}) have commonly served as either source or target task. 
However, among currently existing video datasets, Kinetics has been the standard HAR benchmark~\cite{har-study}, it is commonly used as source task for pretraining and TL studies~\cite{RAY2023100142}, and there are pretrained models available for several video architectures.

Therefore, a common question when performing TL for video analysis is how transferable ImageNet and Kinetics are to other tasks.
In particular, for ImageNet, multiple pieces of research have studied factors that influence its transferability to other similar image tasks~\citep{yosinski,off-the-shelf,taskonomy}.
In the video domain, most works on transferability have focused on HAR to HAR settings~\citep{kataoka2020would,resNet2+1d,hara2018can} and there has been very little research on the transferability from HAR to other video tasks that occur in different settings~\citep{li2020word,singh2021road}.
For instance, the multi-label movie trailer genre classification (MTGC) is a video task of very distinct nature and content in which transferability has barely been studied.
This task is difficult because genres do not have a specific physical expression in a frame or a sequence of frames.
Consequently, genres must be inferred from characters, scenes, themes, dynamics, relationships and other abstract elements.
In addition, the task implies a natural subjectivity: different human observers can assign different genres to the same trailer.
In contrast to HAR clips, trailer settings are more diverse; namely, the story is usually not presented linearly, fictional elements may be included (e.g., characters, landscapes, devices, laws of physics, etc.) and the duration is generally longer.
Hence, there are important dissimilarities in image content, video structure, and duration between IC/HAR and MTGC tasks that could affect the transferability of spatial and spatio-temporal representations learned by pretrained models on ImageNet and Kinetics.

Over the past decade, multiple works have proposed the use of deep neural network architectures to tackle MTGC. 
Although these works have performed TL at some level, they have not explicitly studied the aspects that may affect transferability of spatial and spatio-temporal representations. 
On the other hand, the evaluation of MGTC methods have been carried out on different movie trailer datasets that range from a few thousand trailers and only four genres to tens of thousands of trailers and twenty-eight genres.
These datasets are typically collected by gathering titles and genres from IMDb\footnote{Internet Movie Database: \url{https://www.imdb.com/}} and video trailers are downloaded either from YouTube\footnote{YouTube: \url{https://www.youtube.com/}} through an automatic title search or in some cases directly from IMDb itself when it is available.
However, this process can be prone to errors since some downloaded videos may not correspond to the actual movie title or they may have significant amounts of advertisement and/or padding, which deteriorates the quality of the dataset.
This is particularly relevant for TL, because the transferability of spatial and spatio-temporal representations can be affected by the sample and labeling quality of both source and target datasets~\cite{Zhang2019TransferAL,kataoka2020would,9938381}. 
Accordingly, to study the transferability of such representations on MTGC, a movie trailer dataset with high quality labeling and video samples is essential.

\subsection{Research Objectives and Contributions}
In this paper, we aim to study the transferability of spatial and spatio-temporal representations to frames and sequence of frames of video trailers for the genre classification task. Specifically, our research objectives are as follows.

\begin{enumerate}
    \item To study how transferable representations learned on ImageNet and Kinetics are to MTGC.
    \item To identify important factors that influence transferability and to what extent.
    \item To analyze strategies to reduce structural dissimilarities between source and target tasks that can improve TL.
\end{enumerate}
To address these objectives, this paper makes the following contributions:

\begin{enumerate}
    \item It introduces Trailers12k, a novel video movie trailer dataset composed of 12,000 trailers multi-labeled with ten different genres.
    In order to ensure the labeling and sample quality, in Trailers12k each URL was manually verified and corrected in terms of title-trailer agreement and video quality.
    In addition to metadata, frame-level and clip-level representations obtained with pretrained models on ImageNet and/or Kinetics are provided.
    \item It proposes Dual Image and Video Transformer Architecture (DIViTA), which is an MTGC architecture that leverages representations learned on ImageNet and/or Kinetics.
    DIViTA improves transferability by performing a straightforward adaptation stage that segments the input trailer based on shot transitions.
    Video segment representations are further aggregated with a transformer-based module to generate a trailer-level representation, which is used for genre classification.
    \item It provides an extensive empirical evaluation of the transferability of convolutional and transformer backbones pretrained on ImageNet and/or Kinetics to Trailers12k MTGC. The impact of the segmentation strategy, frame rate, input video extension and spatio-temporal modeling strategy on classification performance is also studied. In addition, the use of lightweight architectures as an alternative to popular heavier architectures is explored.
\end{enumerate}

This paper is organized as follows. \autoref{sec:relwork} reviews related work on transferability from ImageNet and Kinetics, as well as works on MTGC using neural networks.
\autoref{sec:trailers12k} introduces the Trailers12k dataset, as well as its collection procedure and characteristics.
\autoref{sec:transfer} discusses important data and task dissimilarities between ImageNet, Kinetics, and Trailers12k.
A general overview of the proposed multi-label genre classification architecture DIViTA is presented in \autoref{sec:divita}.
\autoref{sec:experimental} describes the experimental setting and evaluation methodology.
Results are discussed in terms of the transferability from ImageNet and/or Kinetics to Trailers12k MTGC in \autoref{sec:results}.
Finally, the conclusions and perspectives for future work are presented in \autoref{sec:conclu}.

\section{Related Work}
\label{sec:relwork}

This section reviews related works on transfer learning from ImageNet and Kinetics to image and video analysis tasks. It also describes movie trailer genre classification methods, focusing on those using pretrained deep neural networks.

\subsection{Transfer Learning from ImageNet and Kinetics}
ImageNet pretraining has been a widespread practice for image classification~\citep{efficientnet,bit}, object detection~\citep{r-cnn,yolo}, and object segmentation models~\citep{r-cnn,fcn}. 
Transfer learning from ImageNet has been successfully applied not only to generalist classification datasets, but also to fine-grained (birds~\citep{efficientnet}, flowers~\citep{bit}, etc.) and even to domain-specific datasets (chest X-rays~\citep{xie2018pre,ke2021chextransfer}, skin lesions~\citep{lopez2017skin}, etc.).
Surprisingly, some of the latter datasets exhibit very different image distribution attributes and classification formulation than ImageNet.

Several studies have focused on evaluating the transferability of the representations learned on ImageNet.
For instance, \citet{off-the-shelf} compared the performance of pretrained convolutional neural networks as feature extractors against hand-crafted feature algorithms on several image analysis tasks, reporting consistent superior results with convolutional neural networks.
Another work~\citep{yosinski} analyzed the degree of layer specialization at different points in deep neural networks with respect to transferability.
Similarly, \citet{kornblith2019better} investigated the relation between pretraining performance and transfer learning performance using several architectures, finding a strong correlation between accuracy in the ImageNet pretraining task and accuracy in the target task for the corresponding pretrained model.
For generalist image datasets, \citet{taskonomy} proposed the use of the transferability between tasks as a framework to characterize their affinity.
On the other hand, \citet{zhou2021convnets} compared the transferability of ImageNet-pretrained Transformers and ConvNets on several image analysis tasks, finding that Transformers exhibited a greater transferability than ConvNets for most of the tasks.

ImageNet has also been widely used to initialize deep video architectures.
For instance, \citet{i3d} introduced I3D, a 2D ConvNet pretrained on ImageNet that is converted into a 3D video architecture by inflating (copying) trained filters along the temporal dimension.
They showed that initializing I3D with ImageNet widely outperforms random parameter initialization for Kinetics as a target dataset.
Other ConvNet architectures have followed I3D's pretraining approach to improve the state of the art on several HAR datasets~\citep{hara2018can,chen2018multi}.
More recently, pretraining on ImageNet has enabled the introduction of transformer architectures~\citep{timesformer,ViViT,swin-3d} for video tasks.

Similarly, multiple studies have analyzed the transferability of Kinetics to other popular action recognition datasets.
For 3D ConvNets, existing studies~\citep{resNet2+1d,hara2018can} have consistently reported that pretraining on Kinetics outperforms random initialization. 
On the other hand, \citet{kataoka2020would} reported that pretraining on large scale video datasets helps improve the performance of 3D ConvNets, although transferability can benefit more from a dataset with a higher labeling quality (Kinetics-700) than simply from a larger video dataset (Moments in Time~\citep{8651343}).
Kinetics pretraining has also been applied to other action recognition settings, such as egocentric actions~\citep{plizzari2022e2}, action recognition from drones~\citep{choi2020shuffle} or actions in the dark~\citep{xu2021arid}.
Other studies have used Kinetics to initialize models for more distant video tasks, including sign language recognition~\citep{li2020word} or autonomous vehicle decision-making~\citep{singh2021road}.
\autoref{tbl:related_work_transfer} summarizes transferability studies for image and video representations.

\begin{table}[tbp]
\caption{Transferability studies for visual representations.}
\begin{adjustbox}{center}
\centering\footnotesize
\begin{tabular}{@{} l l l l@{}}
\toprule
\multirow{2}{*}[-2pt]{Work} & \multirow{2}{*}[-2pt]{Study} & \multicolumn{2}{c}{Dataset/Task}\\
\cmidrule(lr){3-4}
& & {\scriptsize Source} & {\scriptsize Target} \\
\midrule
\citet{off-the-shelf} & 
CNNs vs. hand-crafted features & ImageNet/IC & Multiple/IC\\
\citet{yosinski} &
Layer specialization & ImageNet/IC & ImageNet,Caltech-101/IC\\
\citet{taskonomy} &
Relationship between tasks & ImageNet/IC & Multiple/IC\\
\citet{kornblith2019better} &
Base vs. target task performance & ImageNet/IC & Multiple/IC\\
\citet{kataoka2020would} &
Transferability between HAR datasets & Kinetics-700/HAR & Multiple/HAR\\
\citet{zhou2021convnets} &
ConvNets vs. Transformers & ImageNet/IC & Multiple/IC \\
\bottomrule
\end{tabular}
\end{adjustbox}
\label{tbl:related_work_transfer}
\end{table}

However, to the best of our knowledge, no explicit studies have been performed on the transferability of ImageNet and Kinetics representations to video tasks with longer videos (spatio-temporal dependencies) and more diverse image (spatial) or video (spatio-temporal) information. In this paper, we investigate transferability for a video task of this kind using a novel movie trailer dataset.

\subsection{Movie Trailer Genre Classification}

Movie trailers are a rich and valuable source of information that have been exploited for different tasks, including movie recommendation~\citep{deldjooMMTF14K}, revenue prediction~\citep{AHMAD2020102278}, story understanding~\citep{movienet}, video summarization~\citep{KANNAN2015286,10.1145/3532213.3532306}, actor recognition~\citep{10.1007/978-981-15-7533-4_77}, age-usability rating of trailers~\citep{shafaei2021case}, and affect-based movie genre classification~\citep{YADAV2020106624}.
Notably, the task of genre recognition from video trailers has been studied over the past couple of decades.
The first approaches to tackle this problem were based on features produced by hand-crafted image algorithms~\citep{rasheed2005use,movie-classification-via-scene,huang2012movie}.
These works casted trailer genre classification as a multi-class problem instead of the more natural multi-label formulation and were evaluated on datasets of only a few hundred trailers. 

More recently, multiple studies have leveraged deep neural networks for MTGC with datasets of increasing size.
Many of these studies have tackled MTGC using only visual information from the trailer videos and have commonly exploited pretrained models.
One of the earliest methods based on deep neural networks was proposed by \citet{cnn-motion}, who applied a VGG-like 2D ConvNet architecture trained from scratch to first independently classify frames and then used different aggregation strategies to obtain a global trailer prediction.
To evaluate this approach, \citet{cnn-motion} introduced a dataset called LMTD comprised of 3,500 trailers with four different genres.
In a follow-up work, \citet{Wehrmann2017} proposed a method that leverages a ResNet architecture pretrained on ImageNet and Places-360 to obtain frame-level representations that were aggregated with the CTT module, a Conv1D block that classifies the whole trailer.
However, the above methods process video frames independently with deep neural networks pretrained on image datasets. Therefore, they don't fully exploit spatio-temporal relations locally encoded in the input trailer frames.

Another important body of research has been devoted to developing MTGC methods that exploit multiple sources of information~\citep{moviescope, cimat, behrouzi2022multimodal, bi2022shot}. 
For instance, \citet{moviescope} proposed a multimodal method that uses video, audio, poster, text, and metadata.
For each modality, this method produces a sequence of representations which are aggregated with a fastText-based~\citep{fasttext} module.
For the video modality, the sequence of representations corresponds to frame-level representations generated by a VGG-16 network pretrained on ImageNet.
The trailer global representation is obtained with an attention module that fuses different modal representations.
This method was evaluated on the Moviescope dataset, which is composed of 5,027 trailer videos, text plots, and multiple metadata.
\citet{cimat} extended the architecture by~\citet{moviescope}, replacing the fasText-based multimodal fusion with a transformer module, which improves performance for MTGC.
Multimodal genre classification for whole movies has recently been enabled by the introduction of MovieNet~\citep{movienet}, a dataset for holistic movie understanding that includes complete movies, subtitles, trailers, synopsis, metadata, etc.
This dataset provides several tasks such as genre classification, action recognition, or cinematic style classification.
While MovieNet includes trailers, they are not considered the main data and roughly half of the trailers do not have a one-to-one relationship with movies because several movies have more than one trailer.
Although these works take into account information from multiple modalities to carry out MTGC, they are not exploiting spatio-temporal relations locally encoded in the input trailer frames either, since video frames are also processed independently.

\begin{table}[tbp]
\caption{Summary of existing MTGC methods based on deep learning. The first three methods process multiple video frames independently, while the last two methods process short video clips.}
\begin{adjustbox}{center}
\centering\footnotesize
\begin{tabular}{@{} l l l l l @{}}
\toprule
Work & \makecell{Pretraining Dataset/\\Backbone Architecture} & \makecell{Genre Classification\\Architecture} & Trailer Dataset & \makecell{Processed\\Information}\\
\midrule
\citet{Wehrmann2017} & ImageNet/Inception-v3 & Conv1D & LMTD-9 & Video Frames\\
\citet{moviescope} & ImageNet/VGG-16 & FastText, Attention & Moviescope & Multimodal\\
\citet{cimat} & ImageNet/CNN & Transformer, GMU & Moviescope & Multimodal\\
\citet{yu2021asts} & Sports-1M/C3D-like & BiLSTM, Attention & MovieTrailer-14k & Video Clips\\
\citet{bi2021video} & Kinetics/I3D & C3D-LSTM & LMTD-9 & Video Clips\\
\bottomrule
\end{tabular}
\end{adjustbox}
\label{tbl:related_work_mtgc}
\end{table}

Pretrained 3D ConvNets have also been exploited to analyze trailer videos, aiming to naturally model spatio-temporal relations among frames.
For instance, \citet{yu2021asts} proposed the Attention based Spatio-temporal Sequential (ASTS) framework, which employs a BiLSTM followed by a Self-Attention module to classify a trailer video from multiple clip representations computed with a C3D network~\citep{tran2015learning} pretrained on Sports-1M~\citep{6909619}.
Similarly, the Video Representation Fusion Network (VRFN) architecture~\citep{bi2021video} computes clip representations with an I3D network~\cite{i3d} pretrained on Kinetics, which are aggregated using an LSTM-based module inspired by CNN-RNN~\citep{wang2016cnn}. \autoref{tbl:related_work_mtgc} summarizes existing works that tackle MTGC using pretrained deep neural networks.

Although the above-mentioned methods have relied on transfer learning at some level to carry out MTGC, transferability has not been explicitly studied.
In contrast, this paper studies different factors that influence transferability and analyzes how complementary ImageNet and Kinetics representations are, whether ConvNet or Transformer architectures provide more transferable representations, and what the trade-off between classification performance and computational complexity is.

\section{\emph{Trailers12k} Dataset}
\label{sec:trailers12k}

\emph{Trailers12k} is a novel movie trailer dataset containing 12,000 titles, each associated with a YouTube video trailer, as well as poster and metadata gathered from IMDb.
The collected information for a sample trailer is illustrated in \autoref{fig:trailers12k}.
\autoref{tbl:datasets} compares data provided by Trailers12k with other similar movie trailer datasets used in previous works.
As can be noted, Trailers12k is the second dataset with more samples and, to the best of our knowledge, the only one in which both title-trailer correspondence and video quality are manually verified.
Moreover, in addition to frame-level spatial representations, it provides clip-level spatio-temporal representations obtained with models pretrained on Kinetics and ImageNet-Kinetics.
The overall compilation procedure can be outlined as follows:

\begin{figure}[tb]
    \centering
    \includegraphics[width=0.9\textwidth]{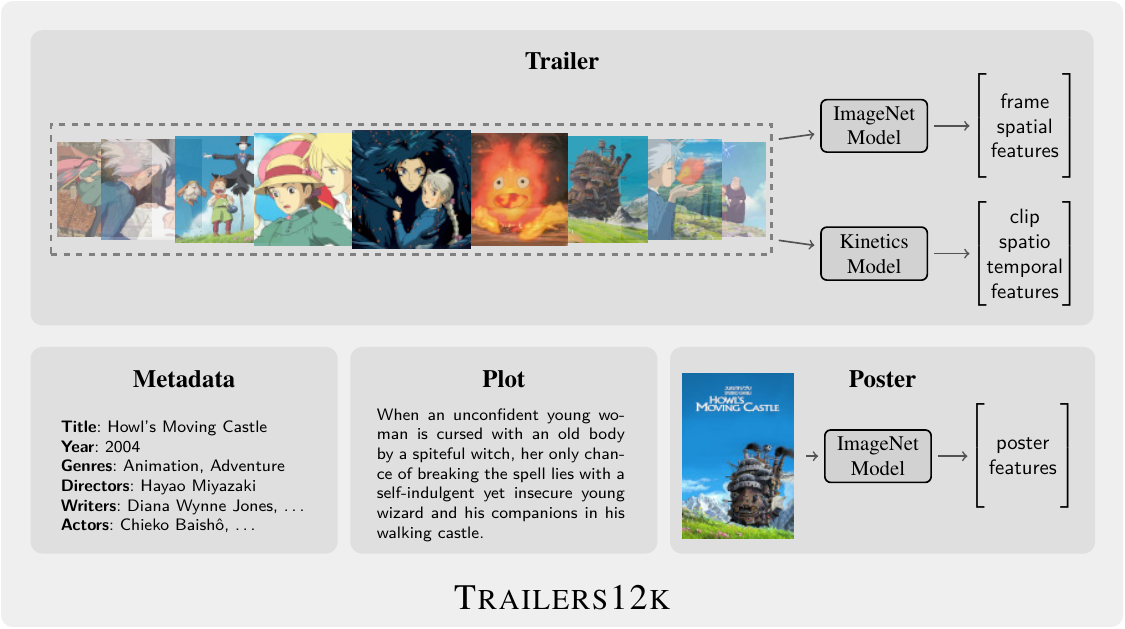}
    \caption{Trailers12k is a high-quality movie trailer dataset comprised of 12,000 titles. It publicly provides metadata, URLs, frame-level and clip-level trailer representations, poster representations, and MTGC evaluation splits.}
    \label{fig:trailers12k}
\end{figure}

\begin{table}[tbp]
\caption{Comparison of Trailers12k to other movie trailer datasets. Columns marked with {\ssmall\faCheck} indicate that data was publicly available to download during the preparation of this table (March 2023). The ImageNet/Kinetics column indicates representations extracted with pretrained models on the corresponding dataset.}
\begin{threeparttable}
\begin{adjustbox}{center}
\centering\footnotesize
\begin{tabular}{@{} l r c r c @{\hspace{1mm}} c @{\hspace{1mm}} c c @{\hspace{1mm}} c c c @{}}
  \toprule
  \multirow{2}{*}[-2pt]{Dataset} & \multirow{2}{*}[-2pt]{Samples} & \multirow{2}{*}[-2pt]{\makecell{Manual\\Verification}} & \multirow{2}{*}[-2pt]{Genres} & \multicolumn{3}{c}{Trailer} & \multicolumn{2}{c}{Poster} & \multirow{2}{*}[-2pt]{Plot} & \multirow{2}{*}[-2pt]{Metadata} \\
  \cmidrule(lr){5-7} \cmidrule(lr){8-9}
  & & & & {\scriptsize URL} & {\scriptsize ImageNet} & {\scriptsize Kinetics} & {\scriptsize URL} & {\scriptsize ImageNet} & \\
  \midrule
  Zhou et al. & 1,239\;\; &
   &  4 & & & & & & & \\
  LMTD & 3,500\;\; &
   &  9 &  &  &  &  &  &  & \\
  MovieScope & 5,027\;\; &
   & 13 & {\ssmall\faCheck} &  &  &  &  &  & \\
  MovieNet & 33,000$^\star$ &
   & 28 &  &  &  & {\ssmall\faCheck} & & {\ssmall\faCheck} & {\ssmall\faCheck}\\
  Trailers12K & 12,000\;\; &
  {\ssmall\faCheck} & 10 & {\ssmall\faCheck} & {\ssmall\faCheck} & {\ssmall\faCheck} & {\ssmall\faCheck} & {\ssmall\faCheck} & {\ssmall\faCheck} & {\ssmall\faCheck} \\
  \bottomrule
\end{tabular}
\end{adjustbox}
\begin{tablenotes}
    \footnotesize
    \item $^\star$MovieNet authors mention 60k trailers in their paper but they correspond to only 33k unique movies.
\end{tablenotes}
\end{threeparttable}
\label{tbl:datasets}
\end{table}

\begin{enumerate}
    \item A list of movie titles released between 2000 and 2019 with the top IMDb user rating was retrieved automatically.
    \item For each movie, an automated YouTube search was performed using the title,  appending the year and the word ``trailer''. The top video result was downloaded.
    \item Titles were filtered to keep only those having at least one of the top 10 most popular IMDb genres and at least 500 user votes.
    \item Since in many cases the resulting trailer did not correspond to the title (it could be a remake, homonym, fan-made trailer, etc.), the trailers were manually curated. Specifically, we manually verified the correspondence of each title-trailer pair and replaced incorrect trailers with the best available on YouTube.
    We also replaced trailers in order to fulfill the following video quality requirements: its duration must be between 60 and 210 seconds, its resolution must be at least 480p, and it should contain the least possible advertising and spatial/temporal padding (color bars/frames). 
\end{enumerate}

\begin{figure}[tb]
    \centering
    \includegraphics[width=0.8\textwidth]{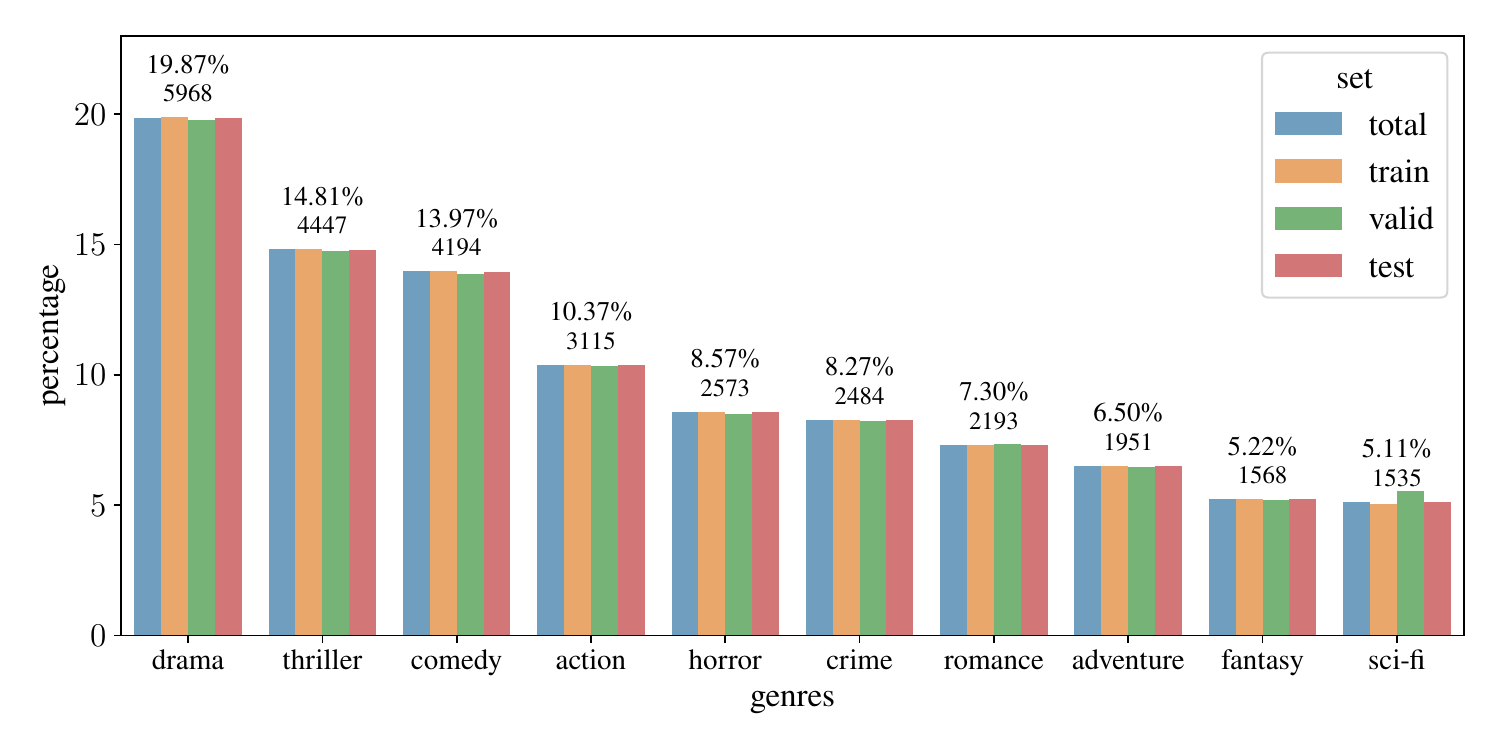}
    \caption{Comparison of the genre distribution in the complete dataset against the distribution in the subsets of the first split. The percentages and sample counts on top of the bars of each genre correspond to the complete dataset (blue).}
    \label{fig:genres_dist}
\end{figure}

In Trailers12k, each title-trailer pair has one or more associated genres, which are indicators of the movie content commonly influencing audience decisions.
In this sense, the dataset is multi-labeled with the top ten most popular IMDb genres: action, adventure, comedy, crime, drama, fantasy, horror, romance, science-fiction, and thriller. 
\autoref{fig:genres_dist} shows the number of examples per genre in the dataset (blue bars). 
A strong genre imbalance can be observed: the most frequent genre (drama) occurs approximately four times more often than the less frequent one (science-fiction).

\begin{figure}[tb]
    \centering
    \includegraphics[width=0.85\textwidth]{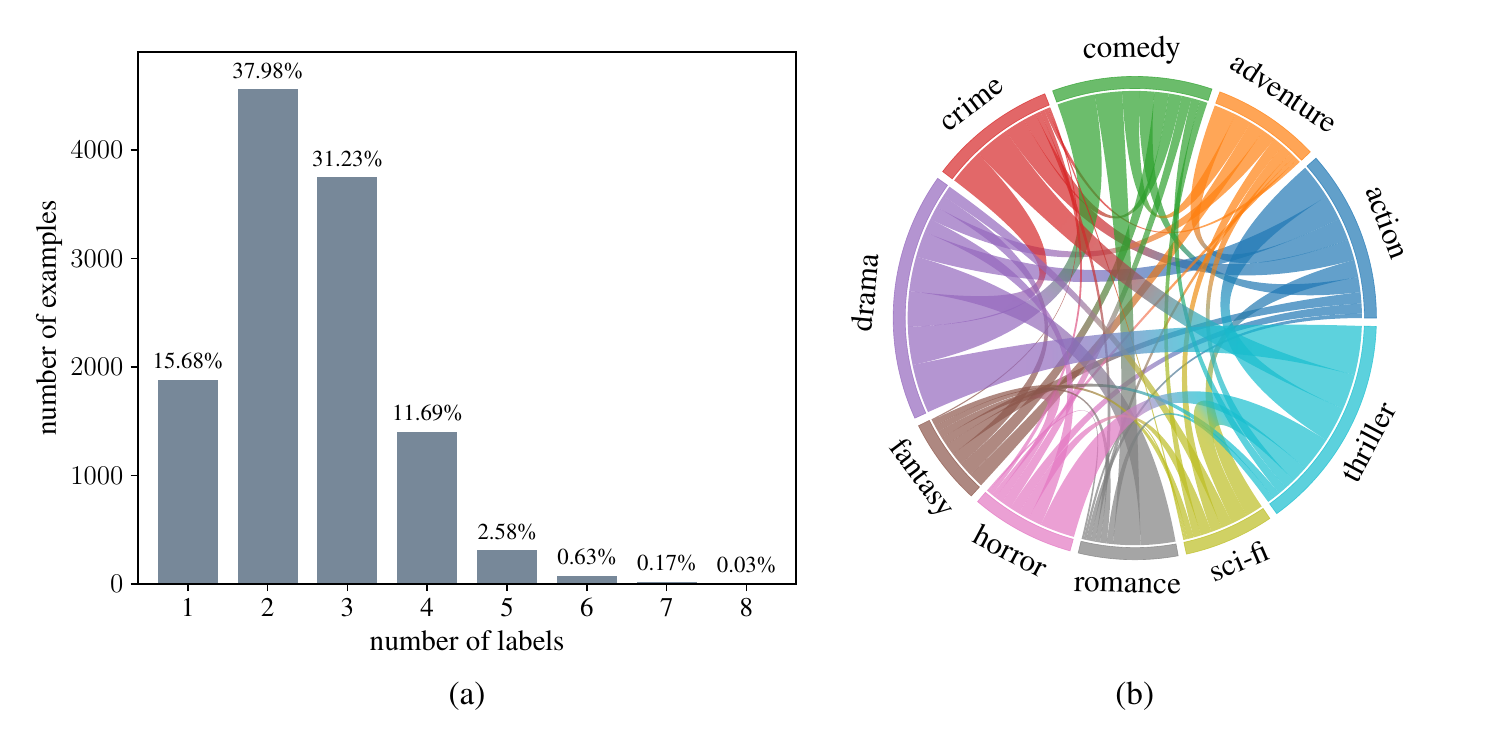}
    \caption{Genre distribution: (a) histogram of the number of labels and (b) correlation between genre pairs.}
    \label{fig:genre_misc}
\end{figure}

The correlation between genres influences other aspects of genre distribution.
To shed light on this, \autoref{fig:genre_misc} (a) shows the histogram of the number of labels.
Note that nearly 70\% of the examples have 2 or 3 labels.
In this sense, the dataset has a label cardinality and density~(see \citet{Tsoumakas2010}) of 2.5 and 0.25, respectively.
Similarly, \autoref{fig:genre_misc} (b) shows a chord correlation diagram between genres. As expected, certain genre pairs like drama-thriller or comedy-romance are quite common, while movies labeled crime-fantasy or adventure-horror are rare.

\begin{figure}[tb]
    \centering
    \includegraphics[width=0.9\textwidth]{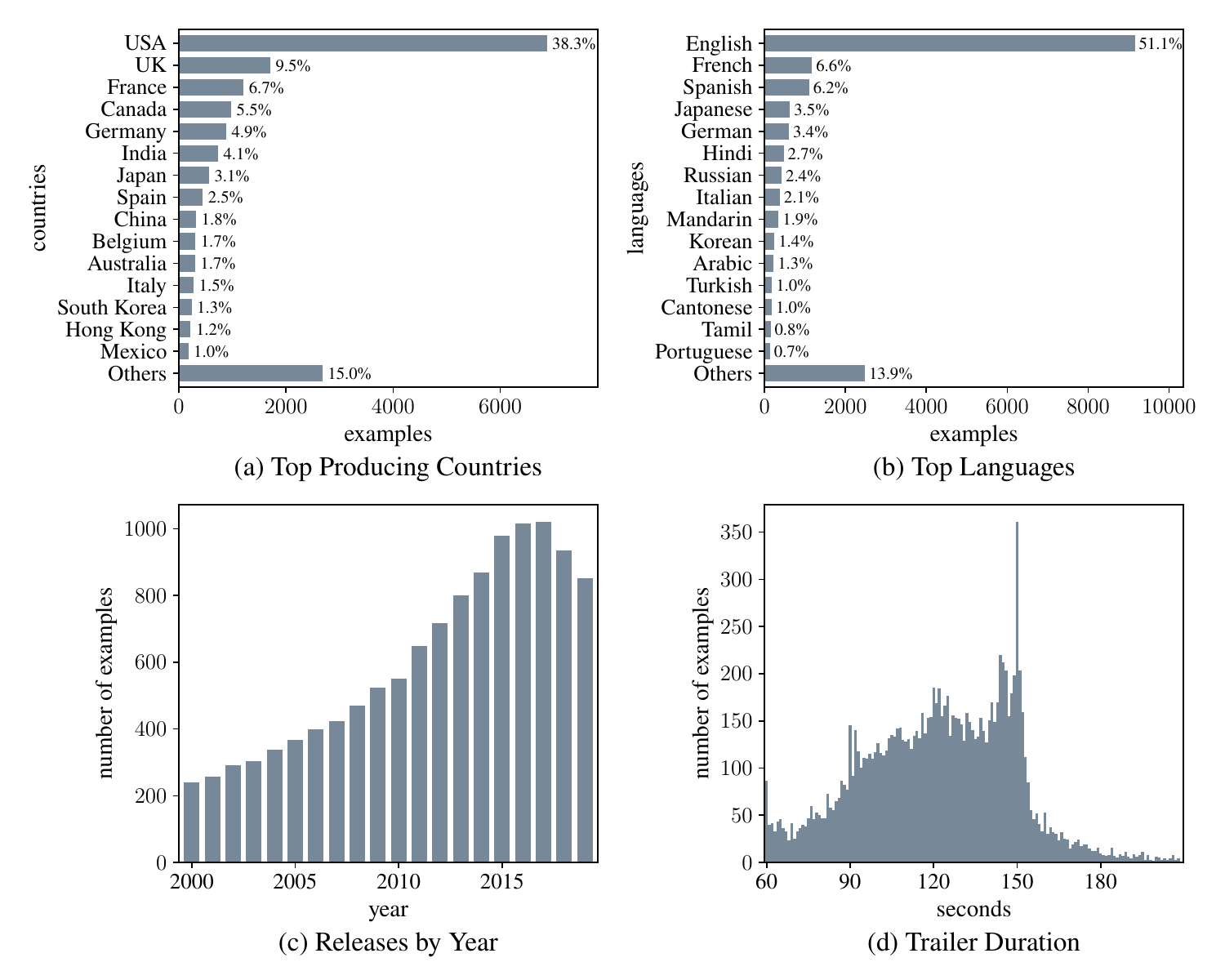}
    \caption{Distribution of different attributes of Trailers12k.}
    \label{fig:atts}
\end{figure}

In addition, each movie title includes metadata gathered from IMDb, such as plots, cast, user rating, number of user votes, languages, synopsis, etc. The metadata attributes and corresponding data types are listed in \autoref{tbl:metadata}.
As shown in \autoref{fig:atts} (a), there are 128 producing countries in Trailers12k, although 85\% of the movies were produced by the top 15 countries.
From a total of 200 spoken languages, 86.1\% of the titles use one of the 15 top languages in \autoref{fig:atts} (b).
As can be seen, USA, UK, and Canada produced 53.3\% of the movies, which explains why English is used in 51.1\% of the titles.
\autoref{fig:atts} (c) illustrates the increasing tendency of movie releases over the years.
Each video trailer has a duration ranging from 30 to 210 seconds, following the distribution in \autoref{fig:atts} (d).
Note that 87.9\% of the trailers have a duration of less than 150 seconds, which corresponds to common industry trailer-making practices~\citep{filmmaking}.
All trailers are normalized to 24 frames per second; in total, Trailers12k contains 407.61 hours of video represented by 35,217,616 frames.

\begin{table}[tbp]
\caption{Trailers12k metadata attributes gathered from IMDb.}
\centering\footnotesize
\def\arraystretch{1.1}
\begin{tabular}{@{} l l @{}}
  \toprule
  Attribute                     & Data type \\
  \midrule
  \texttt{id}                   & string \\
  \texttt{title}                & string \\
  \texttt{year}                 & integer \\
  \texttt{genres}               & list of strings \\
  \texttt{plots}                & list of strings \\
  \texttt{synopsis}             & string \\
  \texttt{cast}                 & list of strings \\
  \texttt{directors}            & list of strings \\
  \texttt{writers}              & list of strings \\
  \texttt{composers}            & list of strings \\
  \texttt{producers}            & list of strings \\
  \texttt{production\_companies}& list of strings \\
  \texttt{languages}            & list of strings \\
  \texttt{certificates}         & list of strings \\
  \texttt{runtime}              & integer \\
  \texttt{votes}                & integer \\
  \texttt{rating}               & float \\
  \texttt{keywords}             & list of strings \\
  \texttt{cover\_url}           & string \\
  \bottomrule
\end{tabular}
\label{tbl:metadata}
\end{table}

As aforementioned, the dataset distribution is influenced by genre imbalance and correlation, which can make model evaluation challenging. To mitigate this difficulty, we provide three different dataset splits, following HMDB~\citep{hmdb51} and UCF101~\citep{ucf101} 3-fold evaluation strategies.
Each split is composed of three subsets, namely training (70\%), validation (10\%), and test (20\%).
To generate the subsets, we used the SOIS~\citep{sois} stratified partition algorithm for multi-label datasets.
This ensures that the generated subsets follow the global genre distribution, as in \autoref{fig:genres_dist} for the first split.
The other two generated splits follow the global distribution in approximately the same way.

Trailers12k YouTube URLs of the trailers, IMDb URLs of the posters, metadata, frame-level and clip-level trailer representations extracted as described in \autoref{sec:divita}, poster representations and evaluation splits are all publicly available at the dataset website\footnote{\scriptsize{\url{https://richardtml.github.io/trailers12k}}}, as well as in Zenodo\footnote{\scriptsize{\url{https://doi.org/10.5281/zenodo.5716409}}}.

\section{Dissimilarities between ImageNet/Kinetics and Trailers12k}
\label{sec:transfer}

The standard approach to perform transfer learning with neural networks can be generally divided into three steps.
First, a model is pretrained on a source dataset.
Second, a new architecture is adapted for the target task which reuses part of the pretrained model.
For target classification tasks, this adaptation step commonly consists of removing specific layers related to the source task (last layers) and replacing them with layers suited for the target task.
Finally, the new architecture is trained on the target dataset.
It has been shown that the performance on the target task is influenced by factors like dataset size~\citep{bit,10.1007/978-3-319-46349-0_5,cherti2021effect}, domain variability~\citep{cherti2021effect}, the capacity of the architecture to learn general representations~\citep{efficientnet}, and the similarity between source and target datasets.
Multiple works~\citep{yosinski,off-the-shelf,taskonomy} have studied transfer learning for different image analysis tasks and have consistently found that a greater similarity between the source and the target tasks results in better transferability, yielding a higher performance on the target task.
In general, a positive transfer occurs when transfer learning benefits the performance on the target task compared with random initialization.
Conversely, if the performance worsens when using transfer learning, it is referred to as negative transfer~\citep{Rosenstein05totransfer}.
Some possible causes of negative transfer that have been identified in the literature include the task/dataset domain divergence, the application of naive transfer methods, and the quality of source and target datasets~\cite{9938381}.

In this paper, we revisit the second step of the transfer learning process and propose a simple adaptation procedure that promotes positive transfer for movie trailer classification.
To understand this procedure, called Snippet Generation stage, let's first analyze important dissimilarities between target and source tasks. 
Although a trailer can be seen as a sequence of correlated images, its content and structure significantly differ from ImageNet images and Kinetics videos, as illustrated in \autoref{fig:dissimilarities}. 
In particular, we focus on the following dissimilarities:

\begin{figure}[tb]
    \centering
    \includegraphics[width=0.9\textwidth]{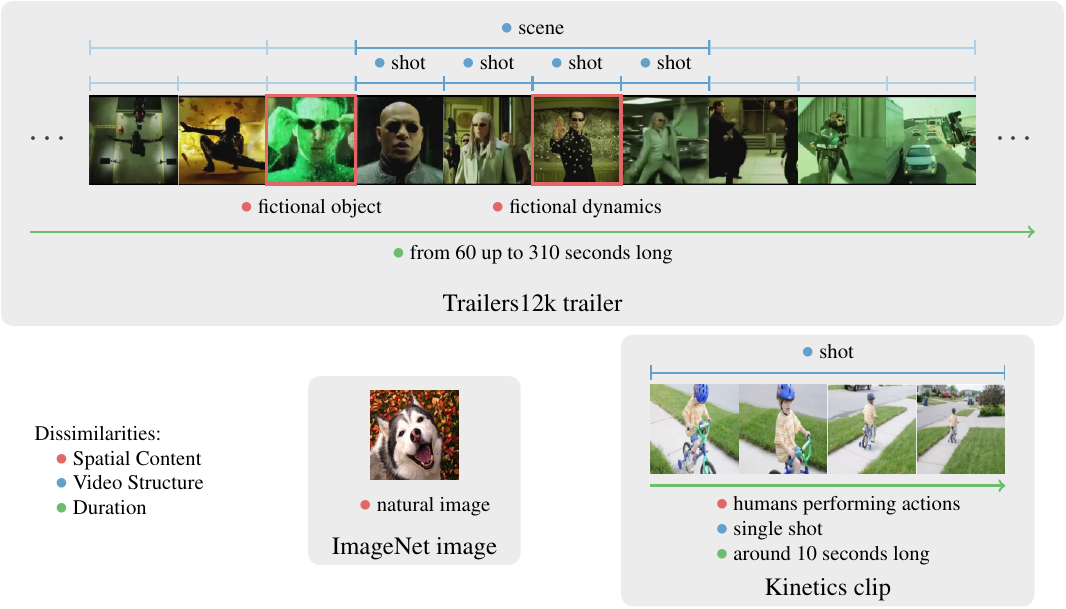}
    \caption{Dissimilarities between ImageNet, Kinetics, and Trailers12k. The common trailer structure is a composition of frames, shots, and scenes presenting content and dynamics of almost any nature.}
    \label{fig:dissimilarities}
\end{figure}

\begin{enumerate}[label=(\alph*)]
    \item \emph{Spatial Content}: It is common for a movie trailer to present fictional elements (e.g., characters, objects, scenes, etc.), actions or dynamics (e.g., violating physical laws) that are not present in ImageNet real-world images or Kinetics human action clips.
    \item \emph{Video Structure}: Since trailers are summaries generated from movies, their spatio-temporal structure is much more complex than HAR clips. Movies use a complex composition for storytelling~\citep{Sklar1993FilmAI}. The most elemental unit is the frame, a still image. A shot is a succession of frames without a camera cut. Generally, a shot has a single background focused on characters or objects that appear in the majority of frames, may be exhibiting some kind of dynamics (e.g., two people hugging). Commonly, a shot lasts from a fraction of a second to a few seconds. Moving up in the film composition, a scene is used to present a narration block through a series of shots with continuity of location, characters, and time. This normally is a few seconds long, but can last up to a few minutes. The filmmaking process has higher compositions, like sequences and acts, but they are used only for movie storytelling.
    As summaries aimed at capturing audience attention over a short period of time, trailers make use mainly of shots and scenes selected from the whole movie. 
    Generally, the chosen shots and scenes are arranged in a sequence that does not usually correspond to the temporal order of the movie.
    The complex composition used by trailers differs considerably from HAR clips that are composed of a few frames focused on humans performing an action.
    \item \emph{Duration}: Trailer12k videos have an average duration ($\approx$ 122s) that exceeds by one order of magnitude the duration of Kinetics-400 clips (10s)~\citep{kinetics400}. For video data analysis, this implies capturing more and longer term temporal information and more computing resources needed for processing.
\end{enumerate}

The Snippet Generation stage aims to improve transfer learning by reducing dissimilarities (b) and (c). This procedure is described in detail in \autoref{sec:divita}.

\section{DIViTA Classification Architecture}
\label{sec:divita}

\begin{figure}[tb]
    \centering
    \begin{adjustbox}{center}
    \includegraphics[width=1\textwidth]{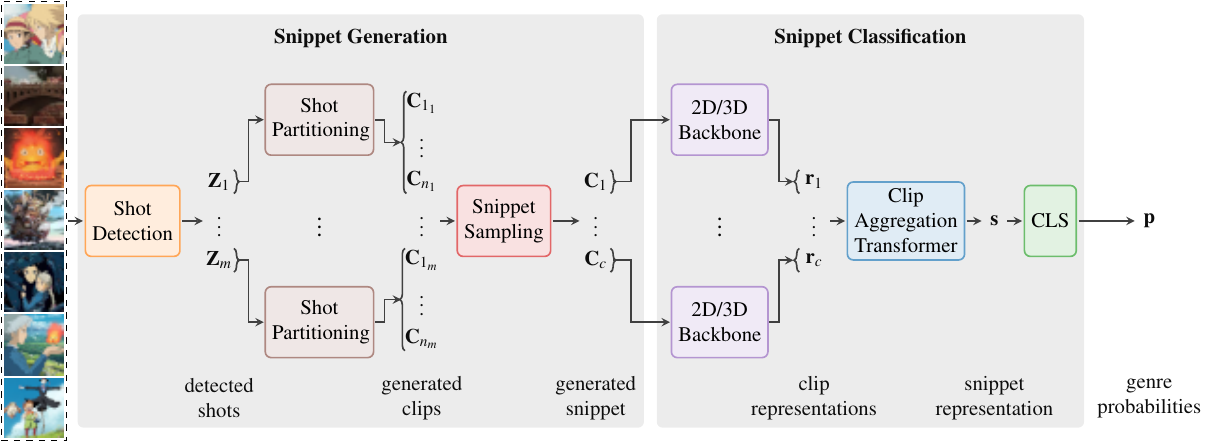}
    \end{adjustbox}
    \caption{Overview of DIViTA's processing steps.}
    \label{fig:divita}
\end{figure}

The proposed classification architecture DIViTA has two general stages, namely Snippet Generation and Snippet Classification, as illustrated in \autoref{fig:divita}.
Roughly speaking, the Snippet Generation stage extracts from the input trailer a short video snippet composed of a sequence of clips, where each clip is preprocessed to become a more suitable input for the pretrained backbone.
The Snippet Classification stage takes the extracted trailer snippet as input, generates a snippet representation by aggregating spatial/spatio-temporal clip-level representations, and classifies the trailer using this representation.

More specifically, the Snippet Generation stage takes an input trailer with $l$ frames and extracts a snippet $\mathbf{S}$ with $c$ clips, each with $f$ frames.
This preprocessing stage is carried out in four steps.
First, a shot detection algorithm partitions the input trailer into a sequence of $m$ shots $(\mathbf{Z}_1, \dots, \mathbf{Z}_m)$, which are short video segments of variable length demarcated by detected transitions (black frames, cuts, fades, etc.).
This step aims to approximate the concept of movie shot used during the trailer production process, as described in \autoref{sec:transfer}.
In the second step, each shot $\mathbf{Z}_i$ is partitioned into smaller segments called trailer clips 
$$
\mathbf{Z}'_i = (\mathbf{C}_{1_i}, \dots, \mathbf{C}_{n_i}),
$$

\noindent where the first $n_i-1$ clips have $f$ frames.
For the last clip $\mathbf{C}_{n_i}$, if it is smaller than $f$ frames, then it is right padded with black frames up to $f$.
At this point, the trailer has been transformed into a sequence composed of the clips of all shots
$$
\mathbf{T} = (\mathbf{C}_{1_1}, \dots, \mathbf{C}_{n_1}, \dots, \mathbf{C}_{1_m}, \dots, \mathbf{C}_{n_m}).
$$ 

\noindent In the third step, a snippet sampling is performed by selecting $c$ adjacent clips from $\mathbf{T}$ to form a trailer snippet $\mathbf{S}$.
Note that these steps generate a video snippet with a high correlation at two different levels, at a lower (inter-frame) level since the frames of a clip belong to the same detected shot, and at a higher (inter-clip) level because all the clips of a snippet are adjacent.
In the last step, which is only performed for a 2D Backbone, each clip $\mathbf{C}_j \in \mathbf{S}$ is represented by selecting a single frame out of the $f$ frames.

\begin{figure}[tb]
    \centering
    \includegraphics[width=0.9\textwidth]{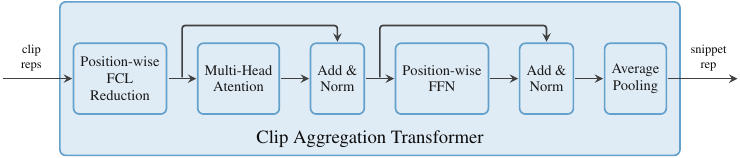}
    \caption{Clip Aggregation Transformer module based on the original Transformer module by~\citet{AttYouNeed}. At the beginning of the module, a Position-wise Fully Connected Layer is incorporated to decrease the size of clip-level vectors.}
    \label{fig:tsfm}
\end{figure}

On the other hand, the Snippet Classification stage is a deep neural network architecture that classifies the preprocessed trailer snippet, and consists of three modules: a 2D/3D Backbone to obtain spatio-temporal representations of trailer clips, a transformer-based module to aggregate spatio-temporal information and a linear layer (CLS) for classification.
The 2D/3D Backbone generates a representation vector $\mathbf{r}_j \in \mathbb{R}^{b}$ for each snippet clip $\mathbf{C}_j$, where $b$ is the backbone output size.
This module is constructed by transferring the representation extraction layers of a classification architecture pretrained on ImageNet and/or Kinetics.
In the case of an image 2D Backbone, $\mathbf{C}_j$ is a single frame, so $\mathbf{r}_j$ encodes purely spatial information.
In contrast, for a video 3D Backbone, since $\mathbf{C}_j$ is a sequence of frames, $\mathbf{r}_j$ encodes spatio-temporal information.
In either case, the output is a sequence of clip representations $(\mathbf{r}_1, \dots, \mathbf{r}_c)$, which is combined by the Clip Aggregation Transformer into a single vector $\mathbf{s} \in \mathbb{R}^{d}$ with spatio-temporal information at the snippet level.
The architecture of the Clip Aggregation Transformer is illustrated in \autoref{fig:tsfm}.
This architecture is based on the original Transformer by~\citet{AttYouNeed}, but incorporates an additional Position-wise Fully Connected Layer at the beginning to reduce each input clip representation $\mathbf{r}_j$ to a vector of size $d < b$.
The intuition behind adding the latter layer is that it can reduce parameter explosion in the following layers and thus helps mitigate overfitting.
The next four blocks in the Clip Aggregation Transformer produce a new sequence of clip representations that is intended to capture dependencies among clip representations.
The Average Pooling layer at the end is applied over the time dimension to aggregate the sequence into a single representation vector $\mathbf{s}$ for the whole snippet.
Finally, CLS is a fully connected layer followed by a sigmoid activation function that classifies $\mathbf{s}$, producing a vector $\mathbf{p}$ of size $g$ where each element represents the probability that the snippet belongs to a given genre.

DIViTA has two operation modes.
At training time, a single snippet $\mathbf{S}$ is used to classify the input trailer.
In the third step of the Snippet Generation, a trailer snippet $\mathbf{S}$ composed of $c$ adjacent clips is selected by picking a starting position from $\left [1, |T| - c \right]$ uniformly at random.
In this operation mode, the snippet probability vector $\mathbf{p}$ is considered to be the classification for the complete trailer.
Since different snippets can be generated from different starting positions, this strategy provides an implicit temporal augmentation effect and at the same time lowers the computational requirements during training.
At inference time, all the snippets $(\mathbf{S}_1, \dots, \mathbf{S}_q)$ are used to classify the input trailer.
In the third step of the Snippet Generation, the trailer clip sequence $\mathbf{T}$ is partitioned into a sequence of snippets $(\mathbf{S}_1, \dots, \mathbf{S}_q)$.
In inference mode, the classification for the complete trailer is obtained by genre-wise averaging the probability vectors $(\mathbf{p}_1, \dots, \mathbf{p}_q)$ of all the snippets.

\section{Experimental Setup}
\label{sec:experimental}

For the empirical evaluation, we fix some training and model hyperparameters while studying the impact of other hyperparameters in terms of different performance metrics. We also compare the performance of our models with baselines. Below, we detail our experimental setup. The code\footnote{\scriptsize{\url{https://github.com/richardtml/DIViTA}}} to reproduce our main results is publicly available.

\paragraph{Training}
Models are trained using the binary cross entropy given by

$$
\mathcal{L}(\mathbf{Y}, \mathbf{P}) = - \frac{1}{mg} \sum_{i=1}^m \sum_{j=1}^g {(y^{(i)}_j \log(p^{(i)}_j) + (1 - y^{(i)}_j) \log(1 - p^{(i)}_j))},
$$

\noindent where $y^{(i)}_j$ and $p^{(i)}_j$ are the ground truth label and prediction probability for the genre $j$ of the example $i$ in the batch, $g$ is the number of genres, and $m$ is the batch size.
Models are trained for 100 epochs in batches of 32 examples. We adopt early stopping based on the validation set loss.
We use the AdamW optimizer with an initial learning rate of 1e-4 which is decreased by a factor of 10 every time the validation loss plateaued for 20 epochs.
Backbone weights are frozen to reduce computing and time resources.
For all experiments, we use a DGX A100 Server.

Unless stated otherwise for a particular experiment, the default configuration for the Snippet Generation stage is set to 30 clips per snippet, each of which is composed of 24 frames taken from the output of the shot detector (Shot-$24$).
The Snippet Classification stage uses a Swin-2D~\citep{swin-2d} backbone pretrained on ImageNet-1K or a Swin-3D~\citep{swin-3d} backbone pretrained on ImageNet-1K and Kinetics-400. The Clip Aggregation Transformer has 4 heads with linear projections of 128 dimensions. Training and architecture hyperparameters for the best configuration of DIViTA are listed in \autoref{tbl:hp}.

\begin{table}[tbp]
\centering
\caption{Hyperparameters for the best DIViTA configuration.}
\begin{adjustbox}{center}
\centering\footnotesize
\begin{tabular}{@{} l c @{}}
\toprule
Parameter & Configuration\\
\midrule
\multicolumn{2}{l}{\hspace{-5pt}\textit{Training}}\\
\hspace{5pt}Weight initializer & Kaiming uniform\\
\hspace{5pt}Optimizer & AdamW\\
\hspace{5pt}Base learning rate & 1e-4\\
\hspace{5pt}Batch size & 32\\
\hspace{5pt}Training epochs & 100 \\
\hspace{5pt}Learning rate decay & 0.1 \\
\hspace{5pt}Learning rate schedule & Reduce on plateau\\
\hspace{5pt}Reduce on plateau patience & 20 epochs\\
\multicolumn{2}{l}{\hspace{-5pt}\textit{Architecture}}\\
\hspace{5pt}Clips per snippet & 30\\
\hspace{5pt}Frames per clip & 24\\
\hspace{5pt}Frames per clip & 24\\
\hspace{5pt}2D Backbone & Swin-2D\\
\hspace{5pt}3D Backbone & Swin-3D\\
\hspace{5pt}Clip Aggregation Transformer & 4 heads of 128\\
\bottomrule
\end{tabular}
\end{adjustbox}
\label{tbl:hp}
\end{table}

\paragraph{Evaluation}
We choose four metrics based on the area under the precision-recall curve commonly used in multi-label trailer classification works~\citep{Wehrmann2017,moviescope}.
$\mu AP$ (micro average) is computed using all labels as a single binary classification task.
This metric provides global information regarding the predictions, allowing more frequent classes to have a greater impact on performance.
In the $mAP$ (macro average) metric, an AUC is computed per class and the results are averaged.
This provides performance information regarding the classes independent of their frequency.
$wAP$ (weighted average) is similar to the $mAP$ metric, but the average is weighted by the frequency of the class.
In contrast to $mAP$, $wAP$ takes into account genre frequency.
Finally, $sAP$ (sample average) computes an AUC per example and the results are averaged.
All the models are trained and evaluated on each of the three Trailers12k splits; the mean and standard deviation taken over the three test sets are reported for each metric.

\paragraph{Baselines}
In order to study the impact of different parts of DIViTA, we perform a straightforward ablation study by replacing key components with simpler alternatives.
We also compare DIViTA with two previous unimodal (video frames) MTGC methods: CTT-MMC-A and fastVideo. 
CTT-MMC-A is one of the CTT variants  proposed by~\citet{Wehrmann2017}, whereas fastVideo is the fastText-based model that processes trailer frames in the approach by~\citet{moviescope}.
For a fair comparison, as opposed to the original work, we use a backbone pretrained only on ImageNet for CTT-MMC-A and omit the additional pretraining on Places360.
In addition to these MTGC methods, we compare DIViTA with TimeSformer~\citep{timesformer}, a Transformer-based video classification architecture that learns spatio-temporal features from a sequence of frame-level patches.
We reproduce these architectures as faithfully as possible from the paper descriptions and train and evaluate them on Trailers12k.

\section{Results, Discussion and Implications}
\label{sec:results}
We evaluate the impact of the clip generation strategy and the clip frame rate on the transferability of ImageNet and/or Kinetics representations.
We analyze Convolutional and Transformer backbones, pretrained on ImageNet and/or Kinetics, comparing both performance and computational requirements.
We also study different snippet lengths and snippet aggregation strategies.

\subsection{Shot Partitioning}

\begin{figure}[tb]
    \centering
    \includegraphics[width=0.9\textwidth]{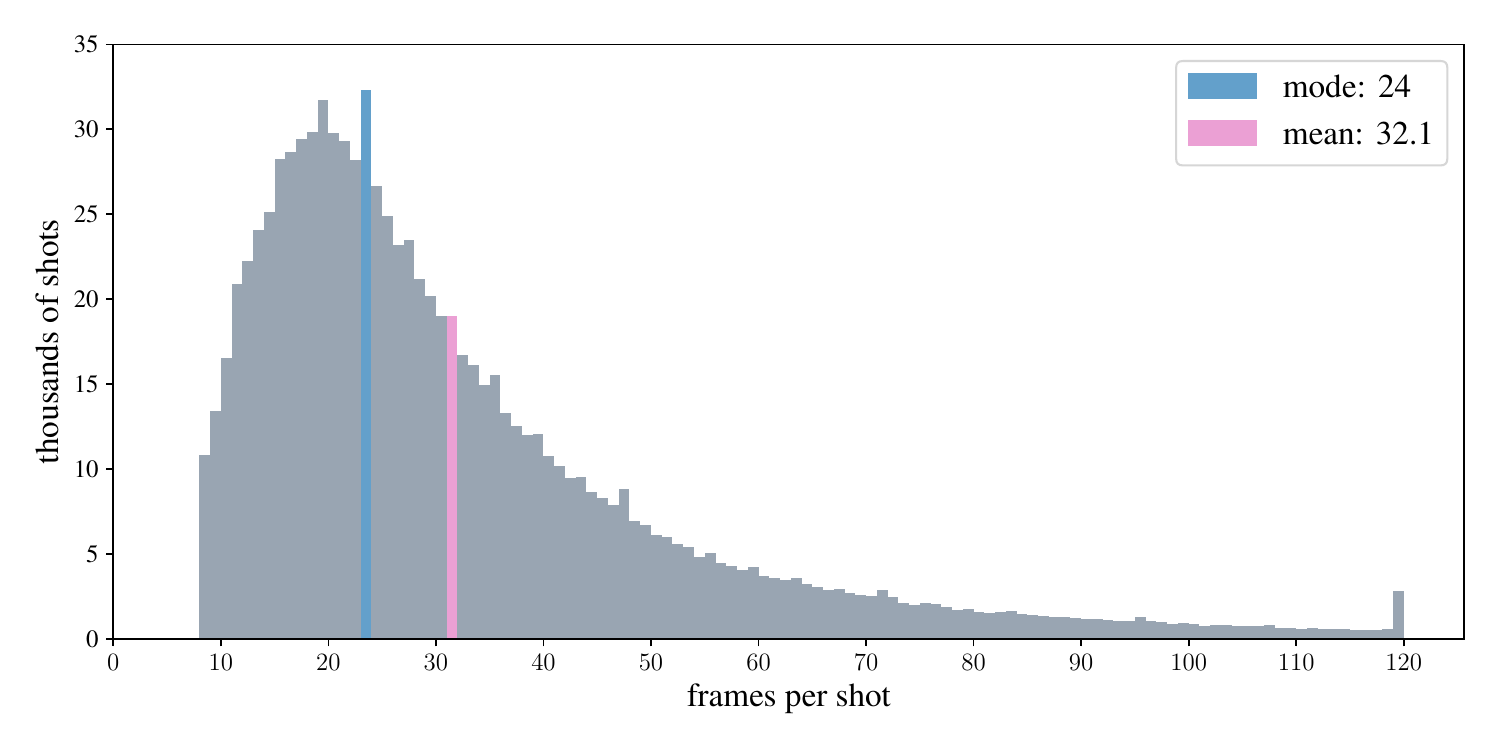}
    \caption{Histogram of shot durations for the Trailers12k  obtained with TransNet~V2~\citep{transnetv2}.}
    \label{fig:shots_hist}
\end{figure}

The first two steps in the Snippet Generation stage are aimed at reducing dissimilarities between images/human action clips and movie shots.
This is achieved by first segmenting the input trailer into shots of variable length, and then partitioning each shot into clips with a fixed number of frames ($f$).
To segment trailers into shots, we use the shot transition detection network TransNet~V2~\citep{transnetv2}, which is an architecture that processes video clips through a stack of blocks based on dilated convolutions (called SDDCNN).
To train TransNet~V2, samples are generated by randomly selecting two video clips and joining them with hard cut or dissolve transitions.
\autoref{fig:shots_hist} shows the histogram of the number of frames per shot for Trailers12k obtained by TransNet~V2.

\begin{figure}[tb]
    \centering
    \includegraphics[width=0.7\textwidth]{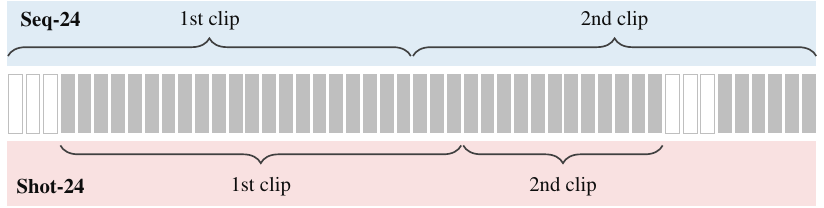}
    \caption{Comparison of clip generation with Seq-24 (blue) and Shot-24 (pink). In the middle, content frames are represented as solid gray rectangles, while transition frames as empty rectangles. In Seq-24, clips are constituted by sequential frames, including both content and transition frames. In Shot-24, clips are made up of sequential content frames only, omitting transition frames.}
    \label{fig:seq_shot}
\end{figure}

Clip length $f$ determines the amount of information provided to the backbone, which may impact transferability.
This is particularly important for Kinetics representations because they are computed with all the clip's $f$ frames.
We explore configurations with $f = 24$ and $f = 32$ clip lengths, called Shot-24 and Shot-32, which correspond respectively to the mode and mean of the distribution of shot durations observed in \autoref{fig:shots_hist}.
To ascertain the benefits of the proposed strategy to generate clips based on shot partitioning, we compare it with a simpler segmentation strategy.
In Seq-24 and Seq-32 configurations, clips are generated by simply taking contiguous sequences of 24 and 32 frames from the trailer.
\autoref{fig:seq_shot} illustrates an example of clip generation with Seq-24 and Shot-24 configurations.

\autoref{tbl:clips_generation} reports results comparing both clip generation strategies Seq-$f$ and Shot-$f$ using ImageNet and ImageNet-Kinetics backbones.
As can be observed, the Shot-$f$ strategy improves transferability, especially for the ImageNet-Kinetics backbone.
More specifically, the highest performance is obtained in all the metrics by Shot-24 using an ImageNet-Kinetics backbone; for instance, it achieves an average $\mu AP$ of 75.57\%, which is 3.75 and 4.1 points higher than Seq-24 and Seq-32, respectively.
Standard deviations are also generally lower for ImageNet-Kinetics backbones with the Shot-$f$ strategy.
This could be an effect of the shot detector, which helps generate clips in which the majority of frames are highly correlated, thus reducing the risk of having transition frames within the clips.
This is particularly important since an ImageNet-Kinetics backbone consumes all the frames of a clip to generate its representation.
For configurations using the ImageNet backbone, the gains are lower; for instance, the Shot-24 $\mu AP$ is around 1.83 points higher than Seq-24.
This can be explained by the fact that a 2D Backbone takes as input a single frame from the clip, selected using the color histogram similarity to the clip's average color histogram, which reduces the probability of taking a transition frame.
For ImageNet-Kinetics, Shot-24 slightly outperforms Shot-32 in all the metrics.
Note that because Shot-24 produces shorter clips, the number of training samples is also 31\% larger than Shot-32.

\begin{table}[tbp]
\centering\footnotesize
\caption{Performance comparison of clip generation strategies Seq-$f$ and Shot-$f$ in DIViTA.}
\begin{tabular}{@{} l @{\hspace{15pt}} c c c c @{}}
  \toprule
  \multirow{2}{*}[-2pt]{\shortstack[l]{Clip Generation}} & \multicolumn{4}{c}{Metrics $\uparrow$} \\
  \cmidrule{2-5}
  & $\mu AP$ & $mAP$ & $wAP$ & $sAP$ \\
  \midrule
  \multicolumn{5}{l}{\hspace{-5pt}\textit{ImageNet}} \\
  \hspace{5pt}Seq-24 &
  70.83±1.93 & 66.39±1.86 & 70.29±1.03 & 76.04±1.83 \\
  \hspace{5pt}Seq-32 & 
  70.13±2.03 & 66.31±2.12 & 70.13±2.05 & 75.95±2.04 \\
  \hspace{5pt}Shot-24 & 
  72.66±1.37 & 67.68±1.36 & \textbf{71.76±1.09} & \textbf{77.49±1.18} \\
  \hspace{5pt}Shot-32 & 
  \textbf{72.90±1.20} & \textbf{67.77±1.58} & 71.70±1.10 & 77.45±1.11 \\
  \multicolumn{5}{l}{\hspace{-5pt}\textit{ImageNet-Kinetics}} \\
  \hspace{5pt}Seq-24 &
  71.82±1.33 & 66.55±1.24 & 69.88±1.61 & 76.01±1.24 \\
  \hspace{5pt}Seq-32 &
  71.42±1.09 & 66.89±1.30 & 69.93±2.04 & 75.94±1.72 \\
  \hspace{5pt}Shot-24&
  \textbf{75.57±0.66} & \textbf{70.48±0.41} & \textbf{74.21±0.40} & \textbf{80.02±0.47} \\
  \hspace{5pt}Shot-32 &
  75.21±0.43 & 69.64±0.48 & 73.32±0.31 & 79.16±0.29 \\
  \bottomrule
\end{tabular}
\label{tbl:clips_generation}
\end{table}

\subsection{Frame Rate}
Movement can be an important clue for video analysis tasks.
For human action clips, the main source of movement comes from humans performing an action and its extent depends on the type of action.
In contrast, in a movie trailer, characters, objects, background or events can independently exhibit great variability in the amount of movement, which also depends on the genre~\citep{movie-classification-via-scene}.
To explore how this aspect impacts transferability, we reduce the frame rate of Trailers12k videos to increase the amount of apparent movement.
The downsampling procedure simply selects frames at equal intervals, e.g., to produce an 8 FPS video only the first out of each three consecutive frames is kept.
\autoref{tbl:frame_freq} reports results at different frame rates.
As we can observe, performance increases as FPS increases, reaching the top result for all the metrics at the original 24 FPS.
Nevertheless, lower frame rates achieve competitive results at a fraction of the computational cost of the top model.
For instance, 4 FPS decreases $\mu AP$ 3.02 points using only $\frac{1}{6}$ of memory to represent the input tensor.

\begin{table}[tbp]
\centering\footnotesize
\caption{Gradual increase of number of frames per clip in DIViTA.}
\begin{tabular}{@{} r @{\hspace{15pt}} c c c c @{}}
  \toprule
  \multirow{2}{*}[-2pt]{\shortstack[l]{Frame Rate \qquad\quad\quad}} & \multicolumn{4}{c}{Metrics $\uparrow$} \\
  \cmidrule{2-5}
  & $\mu AP$ & $mAP$ & $wAP$ & $sAP$ \\
  \midrule
  \multicolumn{5}{l}{\hspace{-5pt}\textit{ImageNet-Kinetics}} \\
  4  & 
  72.55±0.89 & 67.50±0.91 & 71.43±0.86 & 77.40±0.79 \\
  6  & 
  72.93±0.60 & 67.80±0.68 & 71.73±0.71 & 77.89±0.50 \\
  8  & 
  73.04±0.55 & 67.95±0.60 & 71.86±0.57 & 78.24±0.66 \\
  12 & 
  73.34±0.77 & 68.07±0.73 & 71.96±0.39 & 78.63±0.49 \\
  24 &
  \textbf{75.57±0.66} & \textbf{70.48±0.41} & \textbf{74.21±0.40} & \textbf{80.02±0.47} \\
  \bottomrule
\end{tabular}
\label{tbl:frame_freq}
\end{table}

\subsection{Spatio-Temporal Extension}

DIViTA makes use of snippets to loosely approximate movie scenes and use them as shortened representations for the whole trailer.
This simplifies the batch-based training process, implicitly introduces a data augmentation mechanism, and decreases memory and processing requirements.
However, reducing the number of clips per snippet also limits the spatio-temporal receptive field of the Clip Aggregation Transformer module.
Given the weakly supervised labeling nature of Trailers12k genres, this could result in misleading predictions at the snippet level.
Recall that a training snippet is generated by randomly sampling contiguous clips, assigning to it the genres of the whole trailer.
Consequently, if the clips within the snippet do not contain information related to one of the assigned genres, the training process receives a misleading supervisory signal with respect to that genre.
Increasing the number of clips per snippet (approximating the whole trailer) helps alleviate this issue at the cost of reducing the benefits of a shortened representation.
\autoref{tbl:clips_per_snippet} reports results of configurations with increasing spatio-temporal receptive fields.
We observe that the best performance is obtained with snippets of 30 to 40 clips, outperforming even configurations with a larger number of clips.
The performance loss with larger snippets could be a result of reducing the sampling space of snippets during the random selection process.
Interestingly, 10 FPS is just 0.64 $\mu AP$ points below the 30 FPS configuration using only $\frac{1}{3}$ of the memory for the input tensor.

\begin{table}[tbp]
\centering\scriptsize
\caption{Gradual increase of the spatio-temporal receptive field in DIViTA.}
\begin{tabular}{@{} r @{\hspace{15pt}} c c c c @{}}
  \toprule
  \multirow{2}{*}[-2pt]{\shortstack[l]{Clips Per Snippet\qquad}} & \multicolumn{4}{c}{Metrics $\uparrow$} \\
  \cmidrule{2-5}
  & $\mu AP$ & $mAP$ & $wAP$ & $sAP$ \\
  \midrule
  \multicolumn{5}{l}{\hspace{-5pt}\textit{ImageNet-Kinetics}} \\
  5  & 
  72.86±0.69 & 68.78±0.56 & 72.90±0.92 & 77.16±0.98 \\
  10 & 
  74.93±0.77 & 70.25±0.67 & 74.14±1.01 & 79.12±0.38 \\
  15 & 
  75.17±0.69 & 70.36±0.52 & 74.02±0.68 & 79.44±0.45 \\
  20 & 
  75.39±0.65 & 70.42±0.48 & \textbf{74.25±0.59} & 79.69±0.51 \\
  30 & 
  \textbf{75.57±0.66} & 70.48±0.41 & 74.21±0.40 & 80.02±0.47 \\
  40 & 
  75.53±0.68 & \textbf{70.71±0.42} & 74.17±0.39 & \textbf{80.06±0.45} \\
  50 & 
  75.46±0.75 & 70.24±0.40 & 74.02±0.42 & 79.97±0.45 \\
  60 & 
  74.87±0.73 & 70.17±0.53 & 74.02±0.45 & 79.99±0.50 \\
  \bottomrule
\end{tabular}
\label{tbl:clips_per_snippet}
\end{table}

\subsection{Spatio-Temporal Modeling}

The Transformer Clip Aggregation module generates spatio-temporal representations at the snippet level.
From the MTGC task point of view, this loosely captures relations at the trailer scene level.
In order to assess this module's ability to model spatio-temporal relations among clips, we compare it with convolutional and recurrent alternatives.
The main advantage of a transformer architecture lies in the attention mechanism.
This enables capturing relations between any pair of clips, regardless of their relative position within the snippet.
In contrast, a convolutional approach can only find relations among clips within its receptive field determined by filter size.
Similarly, a standard recurrent-based module is limited to model relations sequentially.
We carry out a hyperparameter search to find the best configurations for the convolutional and recurrent modules, preserving those with approximately the same number of parameters.
For the recurrent version, we use one GRU cell with 115 hidden units. 
On the other hand, the convolutional version uses one Conv1D layer with 128 filters of size 3, which is similar to the CTT-MMC-A~\citep{Wehrmann2017} aggregation strategy, but with a different number of filters and filter sizes.

\begin{table}[tbp]
\centering\footnotesize
\caption{Recurrent vs. Convolutional vs. Transformer Clip Aggregation modules.}
\begin{tabular}{@{} l @{\hspace{15pt}} c c c c @{}}
  \toprule
  \multirow{2}{*}[-2pt]{\shortstack[l]{Spatio-Temporal\\Modeling}} & \multicolumn{4}{c}{Metrics $\uparrow$} \\
  \cmidrule{2-5}
  & $\mu AP$ & $mAP$ & $wAP$ & $sAP$ \\
  \midrule
  \multicolumn{5}{l}{\hspace{-5pt}\textit{ImageNet}} \\
  \hspace{5pt}GRU &
  71.43±2.14 & 65.24±1.49 & 69.93±1.59 & 76.22±2.88 \\
  \hspace{5pt}Conv & 
  72.13±2.01 & 66.88±1.44 & 70.89±1.31 & 77.03±2.78 \\
  \hspace{5pt}Transformer & 
  \textbf{72.66±1.37} & \textbf{67.68±1.36} & \textbf{71.76±1.09} & \textbf{77.49±1.18} \\
  \multicolumn{5}{l}{\hspace{-5pt}\textit{ImageNet-Kinetics}} \\
  \hspace{5pt}GRU & 
  74.21±0.97 & 68.28±1.03 & 72.67±0.69 & 78.95±0.66 \\
  \hspace{5pt}Conv & 
  74.40±0.98 & 69.19±1.12 & 73.04±1.02 & 79.30±1.03 \\
  \hspace{5pt}Transformer & 
  \textbf{75.57±0.66} & \textbf{70.48±0.41} & \textbf{74.21±0.40} & \textbf{80.02±0.47} \\
  \bottomrule
\end{tabular}
\label{tbl:cam}
\end{table}

\autoref{tbl:cam} reports results corresponding to the three modules.
In all metrics, the Clip Aggregation Transformer module outperforms the convolutional and recurrent versions for both ImageNet and ImageNet-Kinetics backbones.
In addition, standard deviations are consistently lower.
This can be explained by the attention mechanism's ability to better capture the kind of temporal relations among shots that are common in a trailer, namely, two correlated shots can appear in arbitrary positions within the trailer.

\subsection{ImageNet and Kinetics Transferability}

We use DIViTA to study the degree of transferability from ImageNet and Kinetics to Trailers12k.
We focus on three aspects: pretraining dataset, backbone architecture, and backbone computational requirements.
In our experiments, we consider 
lightweight ConvNets (ShuffleNet-2D~\citep{shufflenetv2-2d} and ShuffleNet-3D~\citep{shufflenetv2-3d}); heavy ConvNets (ResNet~\citep{resnet} and R(2+1)D~\citep{resNet2+1d}); and Vision Transformers (Swin-2D~\citep{swin-2d} and Swin-3D~\citep{swin-3d}).
\autoref{tbl:transfer} summarizes the results of these experiments.

\begin{table}[tbp]
\centering\footnotesize
\caption{Performance comparison of backbone architectures and pretraining tasks, I: ImageNet and K: Kinetics, F: Fusion. Percentages of number of parameters and FLOPS are computed with respect to Swin-3D (top performance).}
\begin{adjustbox}{center}
\centering\footnotesize
\begin{tabular}{@{} l c @{\hspace{4pt}} c c c c c r @{\hspace{4pt}} r r @{\hspace{4pt}} r @{}}
  \toprule
  \multirow{2}{*}[-2pt]{Backbone}
  & \multicolumn{2}{c}{Pretraining}
  & \multicolumn{4}{c}{Metrics $\uparrow$}
  & \multicolumn{2}{c}{Params $\downarrow$} & \multicolumn{2}{c}{FLOPS $\downarrow$} \\
  \cmidrule(lr){2-3} \cmidrule(lr){4-7} \cmidrule(lr){8-9} \cmidrule(l){10-11}
  & I & K 
  & $\mu AP$ & $mAP$ & $wAP$ & $sAP$ 
  & (M) & \% & (G) & \% \\
  \midrule
  \multicolumn{11}{l}{\hspace{-5pt}\textit{Light Conv}} \\
  \hspace{5pt}ShuffleNet-2D
  & {\ssmall\faCheck} & & 
  71.14±0.68 & 66.01±0.46 & 70.17±0.44 & 75.80±0.85
  & 1.7 & 6.09 & 4.3 & 0.27 \\
  \hspace{5pt}ShuffleNet-3D
  & & {\ssmall\faCheck} &
  63.43±1.54 & 58.18±1.50 & 63.59±1.46 & 69.49±1.58
  & 1.7 & 6.09 & 8.5 & 0.53 \\
   \hspace{5pt}ShuffleNet-F
   & {\ssmall\faCheck} & {\ssmall\faCheck} & 
  72.11±0.56 & 67.08±0.37 & 71.42±0.41 & 76.66±0.73
  & 3.3 & 11.82 & 12.9 & 0.81 \\
  \multicolumn{11}{l}{\hspace{-5pt}\textit{Heavy Conv}} \\
  \hspace{5pt}ResNet
  & {\ssmall\faCheck} & &
  71.42±0.59 & 66.63±0.36 & 70.64±0.34 & 76.41±0.38
  & 23.9 & 85.66 & 122.7 & 7.71 \\
  \hspace{5pt}R(2+1)D
  & & {\ssmall\faCheck} &
  70.88±1.37 & 64.99±1.37 & 69.88±1.30 & 75.15±1.08
  & 31.7 & 113.62 & 1823.4 & 114.67 \\
   \hspace{5pt}ResNet-F
   & {\ssmall\faCheck} & {\ssmall\faCheck} &
  73.28±0.66 & 68.03±0.63 & 72.14±0.72 & 77.76±0.44
   & 55.6 & 199.28 & 1946.1 & 122.39 \\
  \multicolumn{11}{l}{\hspace{-5pt}\textit{Transformer}} \\
  \hspace{5pt}Swin-2D
  & {\ssmall\faCheck} & &
  72.66±1.37 & 67.68±1.36 & 71.76±1.09 & 77.49±1.18
  & 27.9 & 100.00 & 114.0 & 7.16 \\
  \hspace{5pt}Swin-3D
  & {\ssmall\faCheck} & {\ssmall\faCheck} &
  \textbf{75.57±0.66} & \textbf{70.48±0.41} & \textbf{74.21±0.40} & \textbf{80.02±0.47}
  & \textbf{27.9} & \textbf{100.00} & \textbf{1590.0} & \textbf{100.00} \\
  \bottomrule
\end{tabular}
\end{adjustbox}
\label{tbl:transfer}
\end{table}

\paragraph{ImageNet vs. Kinetics}
Convolutional backbones pretrained on ImageNet outperform those pretrained on Kinetics.
This is in line with the intuition that it can be easier to predict genres from spatial elements (characters, objects, scenes, etc.) than from dynamic information (actions, motion, etc.).
However, the difference in performance between ResNet and R(2+1)D is only 0.54 $\mu AP$ points.
From a pretraining point of view, this suggests that Kinetics clips (with spatio-temporal information but less spatially diverse) provide a different but competitive knowledge source to ImageNet images (more spatially rich, but purely static).
Moreover, we investigate whether both pretraining datasets can be complementary or not using two different approaches.
The first approach, called Fusion, uses a two-stream workflow inspired by~\citep{two-stream}.
Essentially, it has two identical processing pipelines: one for ImageNet and one for Kinetics, whose prediction logits are averaged (late fusion).
We observe that this approach improves the results with respect to a single pretraining by 0.97 $\mu AP$ points for ShuffleNet-Fusion and 1.86 for ResNet-Fusion.
In the second approach, the transformer backbone architecture Swin-3D is first initialized by inflating, along the temporal dimension, the linear embedding layers of a Swin-2D pretrained on ImageNet~\citep{swin-3d}. 
Then, the initialized Swin-3D architecture is trained on Kinetics.
This dual pretraining outperforms single ImageNet pretraining (Swin-2D) by 2.5 $\mu AP$ points.

\paragraph{ConvNets vs. Transformers}
We compare ResNet and ShuffleNet convolutional architectures to Swin transformer architectures.
We observe that the Swin-3D transformer architecture outperforms convolutional backbones for both ImageNet and/or Kinetics, while having similar standard deviations to the top performance of convolutional backbones.
It is worth noting that Swin-3D achieves the top results using a standard pretraining approach for transformer-based video architectures, in contrast to the Fusion two-stream workflow.
On the other hand, although Swin-2D has higher average performance than 2D ConvNets and ShuffleNet-Fusion, its standard deviation is also higher.

\begin{figure}[tb]
    \centering
    \includegraphics[width=1.0\textwidth]{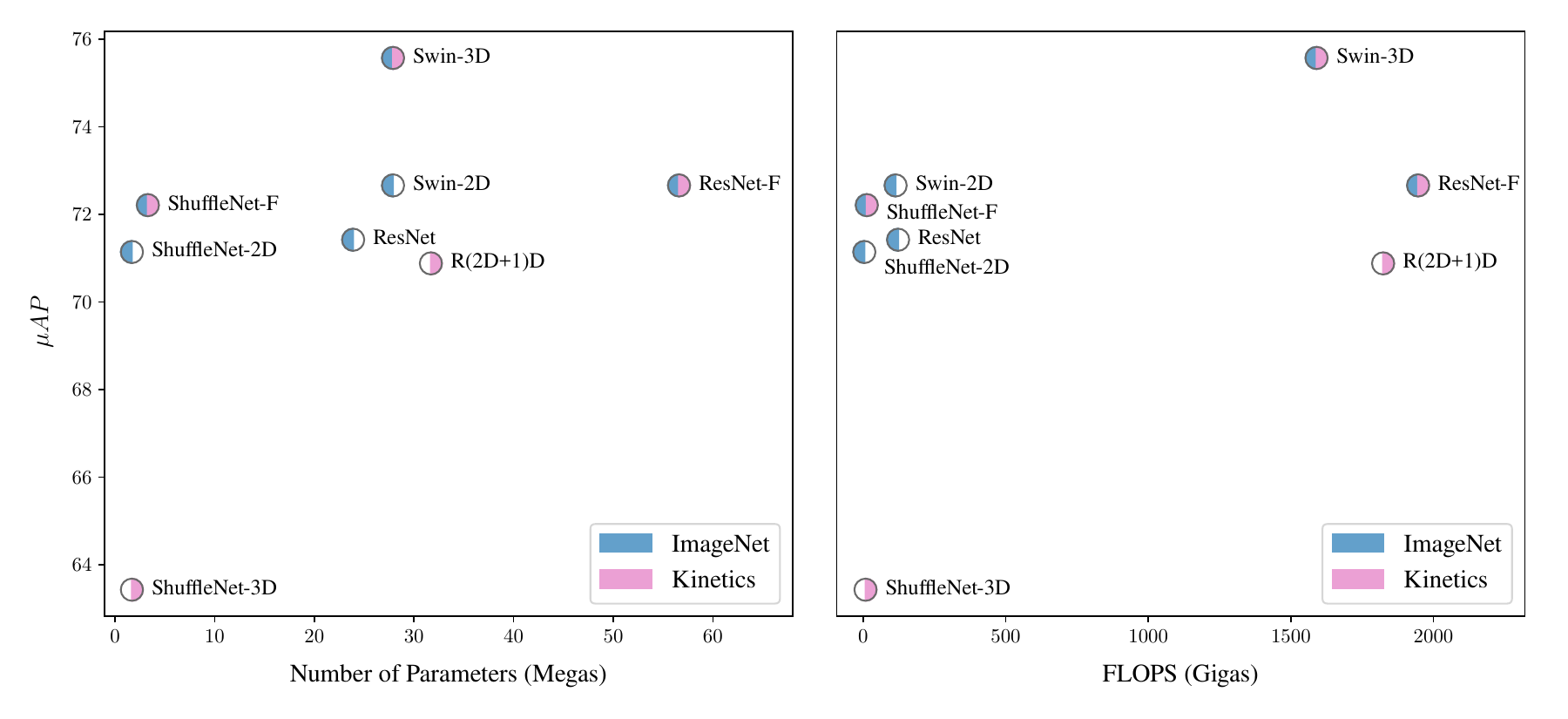}
    \caption{Comparison of different configurations of DIViTA in terms of number of parameters (left) and FLOPs (right) vs. Micro Average Precision $\mu AP$. Models pretrained only on ImageNet are depicted as half-blue bubbles, while those pretrained only on Kinetics as half-pink bubbles. Models that consider both ImageNet and Kinetics pretraining are depicted as half-blue half-pink bubbles.}
    \label{fig:efficiency-performance}
\end{figure}

\paragraph{Computational Requirements}
We analyze the trade-off between performance and computational requirements, as shown in \autoref{fig:efficiency-performance}.
We observe that lightweight ConvNet fusion (ShuffleNet-Fusion) has a performance 3.46 $\mu AP$ lower than the top configuration (Swin-3D).
Nevertheless, the number of parameters is one order of magnitude smaller and the number of FLOPS is two orders of magnitude smaller for ShuffleNet-Fusion.

\subsection{Comparison with Baseline Methods}

In order to assess the overall performance of DIViTA, we compare the configurations with the highest $\mu AP$ to CTT-MMC-A, fastVideo, and TimeSformer. \autoref{tbl:comparison_other} shows the performance of such methods and the DIViTA Swin-2D and Swin-3D configurations.
We can observe that the scores of both DIViTA configurations are higher for all metrics than such baseline methods. 
These results suggest that the combination of the Snippet Generation stage, the Clip Aggregation Transformer and the Swin-3D Transformer backbone can be effective to improve performance when doing TL from ImageNet and/or Kinetics to MTGC.

\begin{table}[tbp]
\centering\footnotesize
\caption{Comparison of DIViTA with baseline methods.}
\begin{adjustbox}{center}
\centering\footnotesize
\begin{tabular}{@{} l c @{\hspace{4pt}} c c c c c @{}}
  \toprule
  \multirow{2}{*}[-2pt]{Method}
  & \multicolumn{2}{c}{Pretraining}
  & \multicolumn{4}{c}{Metrics $\uparrow$} \\
  \cmidrule(lr){2-3} \cmidrule(lr){4-7}
  & I & K 
  & $\mu AP$ & $mAP$ & $wAP$ & $sAP$ \\
  \midrule
  CTT-MMC-A \citep{Wehrmann2017}
  & {\ssmall\faCheck} & & 
  69.27±2.87 & 65.37±1.61 & 68.93±2.09 & 75.09±3.01 \\
  fastVideo~\citep{moviescope}
  & {\ssmall\faCheck} & & 
  68.21±0.73 & 61.19±0.53 & 65.86±0.57 & 74.68±0.68\\
  TimeSformer~\citep{timesformer}
  & {\ssmall\faCheck} & & 
  64.98±1.16 & 59.00±1.07 & 63.26±0.92 & 70.77±0.94\\
  DIViTA Swin-2D (ours)
  & {\ssmall\faCheck} & &
  72.66±1.37 & 67.68±1.36 & 71.76±1.09 & 77.49±1.18 \\
  DIViTA Swin-3D (ours)
  & {\ssmall\faCheck} & {\ssmall\faCheck} &
  \textbf{75.57±0.66} & \textbf{70.48±0.41} & \textbf{74.21±0.40} & \textbf{80.02±0.47} \\
  \bottomrule
\end{tabular}
\end{adjustbox}
\label{tbl:comparison_other}
\end{table}

\subsection{Implications}
The above experimental results have potential implications for performance, efficiency and transferability, namely:

\begin{itemize} 
\item Achieving competitive results with low frame rates is interesting from the video analysis point of view, since it indicates that the temporal information can be substantially downsampled without a significant loss in classification performance, even when using neural network architectures for video.
This finding is consistent with the work by~\citet{X3D} who reported that 3D ConvNets can be applied with low frame rates to HAR in Kinetics, greatly reducing the computational complexity while suffering little performance drops.

\item ImageNet and Kinetics can provide complementary information that benefits transferability for MTGC. However, a higher performance is observed with backbone models pretrained on ImageNet than with backbone models pretrained on Kinetics and just a slightly lower performance than with backbone models pretrained on both datasets.
This suggests that although MTGC is a video analysis task, solely using spatial information could be enough to achieve competitive models for the task.
Recent works studying the importance of temporal information on video tasks have found similar results for certain action recognition classes~\citep{huang2018makes} and video-language understanding tasks~\citep{buch2022revisiting}.

\item In a previous study, \citet{zhou2021convnets} found that representations learned by 2D Transformers on ImageNet outperform those of 2D ConvNets on image analysis target tasks.
Our results extend such transferability findings to ImageNet/Kinetics representations and to a target video task of a more diverse image/video nature.

\item ShuffleNet-Fusion offers an excellent trade-off between performance and computational requirements, which makes it appealing for environments with low computational resources, especially since convolutions have been optimized at both hardware and software levels for these environments~\citep{jetson}.
\end{itemize}

\section{Conclusions}
\label{sec:conclu}

In this paper, we study the transferability of representations learned from generalist image classification and video human action recognition datasets to multi-label movie trailer genre classification.
Specifically, we collected a movie trailer dataset with manually verified title-trailer pairs, which we called Trailers12k, and performed an empirical study of transferability from ImageNet and Kinetics to such dataset.
We also proposed DIViTA, a classification architecture that performs shot detection to segment the trailer into highly correlated clips, providing a more cohesive input for pretrained backbones.
This reduces the gap between the spatio-temporal structure of the source and target datasets, thus improving the transferability of the learned representations.

Our results show that ImageNet and Kinetics representations are comparatively transferable to the MTGC task (e.g. 71.42\% $\mu AP$ for ResNet vs. 70.88\% for R(2+1D)).
Moreover, these representations provide complementary information that can be combined to improve classification performance (e.g. an increase of 1.86\% $\mu AP$ for ResNet-Fusion with respect to ResNet).
In fact, the highest classification performance is achieved with Transformer backbones that leverage both datasets to learn spatio-temporal representations (75.57\% $\mu AP$ for Swin-3D).
Nevertheless, competitive performance with much lower computational requirements can be achieved by separately pretraining dataset-specific ShuffleNets (72.11\% $\mu AP$ for ShuffleNet-Fusion), a family of lightweight convolutional architectures.
Similarly, the computational cost can be decreased by reducing the video frame rate and/or snippet size, greatly lowering the memory and processing requirements while maintaining competitive performance (e.g. 72.55\% $\mu AP$ for 4 frames per clip vs. 75.57\% for 24 frames per clip).
Hence, such configurations are attractive for inference on low-resource environments.
Interestingly, the performance obtained by ImageNet backbones representing clips with a single frame is higher than the performance of Kinetics models and is only slightly lower than ImageNet-Kinetics models, which require multiple frames per clip.

Although our results show that DIViTA is able to effectively transfer representations from ImageNet and Kinetics to MTGC, a possible drawback is that both efficiency and classification performance are affected by the shot detector, which can be computationally expensive and/or inaccurate. 
In particular, our results suggest that using a more accurate shot detector (e.g. Shot-24 or Shot-32) could be computationally expensive, while using a simpler segmentation strategy that takes contiguous sequences of frames (e.g. Seq-24 or Seq-32) could lead to a worse classification performance.
Therefore, in DIViTA we observe a trade-off between efficiency and classification performance. 
On the other hand, DIViTA's design aims to directly capture spatio-temporal relations among the frames of a clip by processing each clip with a 3D ConvNet and Video Transformer backbone, while spatio-temporal relations among the clips of a snippet are modeled more superficially by the Clip Aggregation Transformer.
However, extending DIViTA to more directly capture spatio-temporal relations among the clips of a snippet could be significantly more expensive in terms of FLOPS and memory.

While this piece of research analyzed some factors influencing the transferability of learned representations from two popular computer vision datasets to genre trailer classification, its results have given rise to further questions that can be explored as future research.
For instance, foundation models pretrain architectures on diverse datasets and data modalities, aiming to learn general representations that are transferable to a broader range of tasks.
For genres that contain spatio-temporal information that differs from the representations learned on ImageNet and Kinetics (like fantasy or science-fiction), applying foundation models could be a promising approach to improve results.
In addition, our results suggest that although MTGC could benefit from both ImageNet and Kinetics pretraining, competitive performance at a fraction of model complexity can be achieved using only spatial information.
Therefore, another relevant future research direction is to design architectures and training methods that can model spatio-temporal dependencies more efficiently; e.g., models that simultaneously leverage ImageNet and Kinetics, privileging spatial information while, at the same time, making a more efficient use of spatio-temporal information.
Since the focus of this work was to analyze the transferability of ImageNet and Kinetics representations, DIViTA's design does not use other sources of information from trailers, such as audio, text, posters, or metadata.
However, DIViTA can easily exploit multimodal information by using other representations obtained with backbones for different modalities on the clips and combining the clip representations with a fusion strategy, similar to how spatial and spatio-temporal representations are aggregated in ResNet-Fusion and ShuffleNet-Fusion.
Alternatively, DIViTA can be integrated into multimodal methods, such as Moviescope, by adding it as one of the modality-specific genre models of multimodal fusion strategies.

\section*{Acknowledgements}
This work was supported by a PAPIIT grant [IA104016].
The first author has been supported by the National Council for Science and Technology (CONACYT), Mexico, scholarship number 326014.
We acknowledge the high-performance computing (HPC) resources and services provided by the Corporación Ecuatoriana para el Desarrollo de la Investigación y la Academia (CEDIA).

\bibliographystyle{model5-names}
\biboptions{authoryear}
\bibliography{references}

\end{document}